\documentclass[final]{cvpr}

\usepackage{times}
\usepackage{epsfig}
\usepackage{graphicx}
\usepackage{amsmath}
\usepackage{color}
\usepackage{amssymb,booktabs,tabulary,multirow}
\usepackage[caption=false]{subfig}
\newcolumntype{x}[1]{>{\centering\arraybackslash}p{#1pt}}
\newlength\savewidth\newcommand\shline{\noalign{\global\savewidth\arrayrulewidth
  \global\arrayrulewidth 0.2pt}\hline\noalign{\global\arrayrulewidth\savewidth}}
\newcommand{\tablestyle}[2]{\setlength{\tabcolsep}{#1}\renewcommand{\arraystretch}{#2}\centering\footnotesize}
\makeatletter\renewcommand\paragraph{\@startsection{paragraph}{4}{\z@}
  {.5em \@plus.2ex \@minus.2ex}{-.5em}{\normalfont\normalsize\bfseries}}\makeatother
\usepackage[pagebackref=true,breaklinks=true,colorlinks,bookmarks=false]{hyperref}

\def\cvprPaperID{3082} % *** Enter the CVPR Paper ID here
\def\confYear{CVPR 2021}

\makeatletter\renewcommand\paragraph{\@startsection{paragraph}{4}{\z@}
  {.5em \@plus1ex \@minus.2ex}{-.5em}{\normalfont\normalsize\bfseries}}\makeatother

\begin{document}

%%%%%%%%% TITLE
\title{Points as Queries: Weakly Semi-supervised Object Detection by Points}

% \author{
% Liangyu Chen\\
% Megvii\\
% \and
% Tong Yang\\
% \\

% \author{
%   Liangyu Chen\\
% }

% \author[1]{ABC}
% \auther[2]{Bob}
% \affil[1]{Fudan}
% \affil[2]{Megvii}

\author{Liangyu Chen$^{1,2}$ \thanks{Equally contribution.} \hspace{20pt} Tong Yang$^1$ \footnotemark[1] \hspace{20pt} Xiangyu Zhang$^1$\hspace{20pt} Wei Zhang$^2$ \thanks{Corresponding author.}\hspace{20pt} Jian Sun$^1$\\ 
{$^1 $} MEGVII Technology \hspace{20pt} {$^2 $} Fudan University \\
{\tt\small \{chenliangyu,yangtong,zhangxiangyu,sunjian\}@megvii.com \hspace{20pt} weizh@fudan.edu.cn} 
}

% \author{
% Liangyu Chen\\
% \and
% Tong Yang\\
% \and
% Xiangyu Zhang\\
% \and
% Wei Zhang\\
% \and
% Jian Sun\\
% }
% \institute{MEGVII Technology}

% \auther{
% MEGVII Technology\\
% \and
% Fudan University\\
% }

\maketitle

%%%%%%%%% ABSTRACT
\begin{abstract}
   %The ABSTRACT is to be in fully-justified italicized text, at the top
   %of the left-hand column, below the author and affiliation
   %information. Use the word ``Abstract'' as the title, in 12-point
   %Times, boldface type, centered relative to the column, initially
   %capitalized. The abstract is to be in 10-point, single-spaced type.
   %Leave two blank lines after the Abstract, then begin the main text.
   %Look at previous CVPR abstracts to get a feel for style and length.
    We propose a novel point annotated setting for the weakly semi-supervised object detection task, in which the dataset comprises small fully annotated images and large weakly annotated images by points. It achieves a balance between tremendous annotation burden and detection performance. Based on this setting, we analyze existing detectors and find that these detectors have difficulty in fully exploiting the power of the annotated points. To solve this, we introduce a new detector, Point DETR, which extends DETR by adding a point encoder. Extensive experiments conducted on MS-COCO dataset in various data settings show the effectiveness of our method. In particular, when using 20\% fully labeled data from COCO, our detector achieves a promising performance, 33.3 AP, which outperforms a strong baseline (FCOS) by 2.0 AP, and we demonstrate the point annotations bring over 10 points in various AR metrics. 
    % {\color{red}The code will be available at \url{http://github.com/megvii-model/PointDETR}.}
\end{abstract}

\section{Introduction}
Object detection is one of the fundamental problems in computer vision. Modern object detectors~\cite{law2018cornernet, lin2017feature, lin2017focal, ren2015faster, tian2019fcos} have achieved great success with the help of tremendous annotated data. However, it is very costly to annotate a large amount of detection data. Specifically, for each object instance, a precise bounding box needs to be labeled manually and carefully, which is quite time-consuming: it takes 10-35 seconds~\cite{su2012crowdsourcing,russakovsky2015best,bearman2016s} for labeling an object.

\begin{figure*}[t]
% \begin{subfigure}[b]{.49\linewidth}
% \centering
% \includegraphics[width=1.\textwidth]{01.a.WSOD-example.pdf} 
% \caption{Weakly supervised object detection}\label{fig.examples.WSOD-example}
% \end{subfigure}
% \begin{subfigure}[b]{.49\linewidth}
% \centering
% \includegraphics[width=1.\textwidth]{01.b.SSOD-example.pdf} 
% \caption{Semi-supervised object detection}\label{fig.examples.SSOD-example}
% \end{subfigure}
% \begin{subfigure}[b]{.49\linewidth}
% \centering
% \includegraphics[width=1.\textwidth]{01.c.WSSOD-example.pdf} 
% \caption{Weakly semi-supervised object detection with image-level label}\label{fig.examples.WSSOD-example}
% \end{subfigure}
% \begin{subfigure}[b]{.49\linewidth}
% \centering
% \includegraphics[width=1.\textwidth]{01.d.WSSOD-example-ours.pdf}
% \caption{Weakly semi-supervised object detection with points ({\bf ours})}\label{fig.examples.WSSOD-example-ours}
% \end{subfigure}
\centering
\includegraphics[width=0.96  \textwidth]{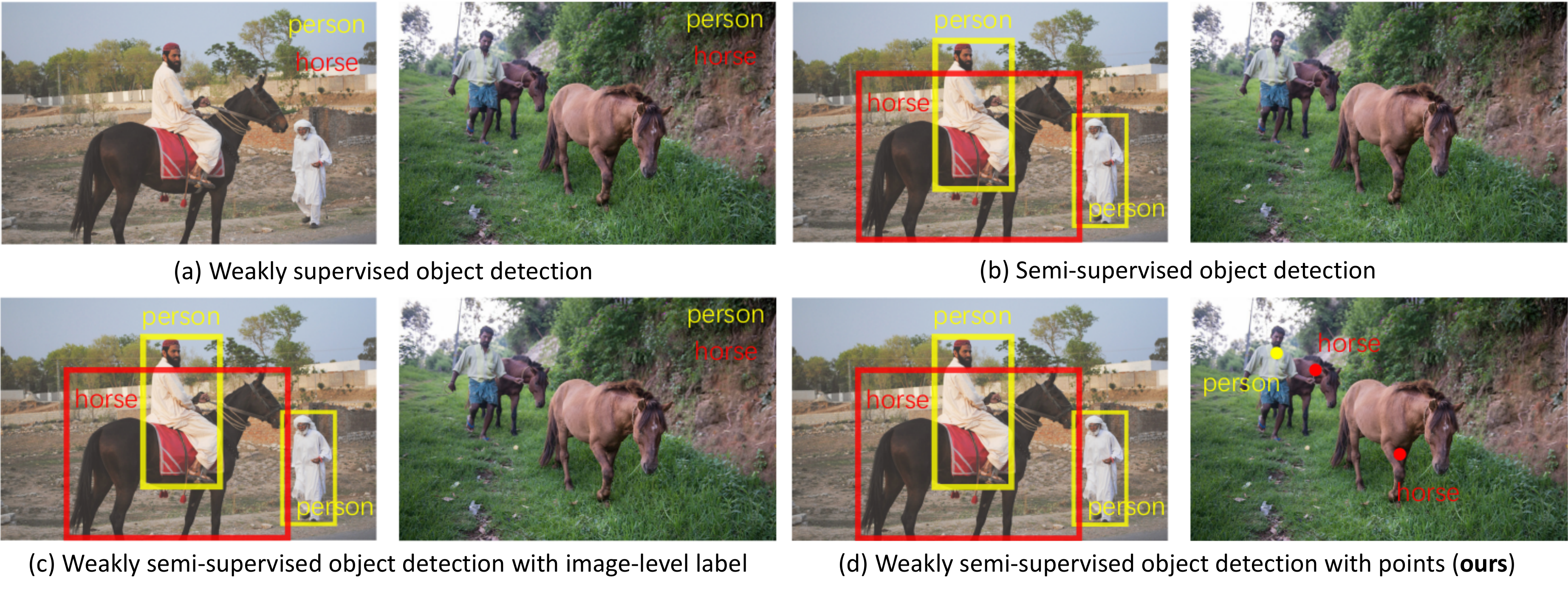}
\caption{Different types of object detection settings to reduce the cost of data annotation}
\label{fig.examples}\vspace{-0.5cm}
\end{figure*}

To reduce the cost of data annotation, weakly supervised object detection (WSOD) and 
% (WSOD, as shown in Figure~\ref{fig.examples.WSOD-example}), 
semi-supervised object detection (SSOD)
% (SSOD, as shown in Figure~\ref{fig.examples.SSOD-example}) 
% and weakly semi-supervised object detection (WSSOD)
% (WSSOD, as shown in Figure~\ref{fig.examples.WSSOD-example})
methods are proposed. Weakly supervised object detection methods~\cite{bilen2016weakly,jie2017deep,shi2017weakly,zhu2017soft} utilize large data with weak annotations, such as image labels, which is far easier to collect than precisely annotated bounding boxes. The semi-supervised object detection methods ~\cite{jeong2019consistency,nguyen2019semi,sohn2020simple,tang2020proposal,wang2018towards} learn detectors with a small amount of box-level labeled images and large unlabeled images, where the cost of image annotation is small. Although these methods can reduce the cost of annotation significantly, their performance is far inferior to their supervised counterparts~\cite{lin2017feature, lin2017focal, tian2019fcos}. To make a trade-off between annotation cost and performance, weakly semi-supervised object detection methods (WSSOD)~\cite{yan2017weakly} are studied, which use small box-level labeled images as well as large weakly labeled images to learn detectors. However, image-level annotations in weakly annotated data are not optimal for object detection task since image labels do not contain the instance-level information of all objects. Motivated by ~\cite{bearman2016s}, we annotate each instance in the image by one point (as shown in Figure~\ref{fig.examples}d) instead of image-level annotation, for two main reasons. Firstly, compared with image-level annotation, points bring much richer information, not only the category of the object but also the strong prior of object location. Secondly, there is no strict requirement on point annotations, such as center points of objects. Thus, the increase in the cost of labeling is marginal compared with the image-level annotation~\cite{bearman2016s}: 23.3 sec/image vs. 20.0 sec/image in VOC dataset~\cite{everingham2010pascal}.
% Therefore, in this paper, we use a small number of fully supervised data and a large number of weakly supervised data labeled by points as our setting based on the above reasons.

Though the above new setting is better for weakly semi-supervised object detection, most recent detectors~\cite{lin2017feature, lin2017focal, tian2019fcos} have difficulty in predicting object boxes based on point annotations. In most detectors, FPN~\cite{lin2017feature} is a basic component, which utilizes multi-level feature maps to predict object boxes. FPN can boost the performance of detectors, but it is incompetent to predict object boxes using point annotations since it is difficult to select the optimal box prediction from multi-level ones, predicted for a point annotation. For the single-level feature detectors, they may suffer from bad performance~\cite{redmon2016you, redmon2017yolo9000, ren2015faster} or strict requirement on point annotations ~\cite{duan2019centernet,law2018cornernet, zhou2019objects} even though they avoid choosing feature map levels.
% To predict bounding boxes from points accurately in our setting, 
% we analyze modern detectors~[xxx, xxx, xxx]. FPN~\cite{lin2017feature} 
% is now a basic component of modern detectors, it makes the detector 
% generates predictions on multi-level feature maps. Although FPN helps 
% to improve the performance of the object detector, it is harmful in 
% the task of predict bounding boxes from points, as not only the points 
% need to be used to predict the accurate bounding boxes, but the correct 
% one needs to be selected from multi-level predictions (details in Section 4.x). 
% Moreover, simply remove FPN in the detector and predicts bounding boxes on a 
% single-level feature map leads to poor performance (details in Section 4.x).

% \begin{figure*}[t]
% \centering
% \subfigure[Weakly supervised object detection]{
% \label{fig.examples.WSOD-example}
% \includegraphics[width=0.47\textwidth]{01.a.WSOD-example.pdf} 
% }
% \subfigure[Semi-supervised object detection]{
% \label{fig.examples.SSOD-example}
% \includegraphics[width=0.47\textwidth]{01.b.SSOD-example.pdf} 
% }
% \subfigure[Weakly semi-supervised object detection with image-level label]{
% \label{fig.examples.WSSOD-example}
% \includegraphics[width=0.47\textwidth]{01.c.WSSOD-example.pdf}    
% }
% \subfigure[Weakly semi-supervised object detection with points ({\bf ours})]{
% \label{fig.examples.WSSOD-example-ours}
% \includegraphics[width=0.47\textwidth]{01.d.WSSOD-example-ours.pdf}
% }
% \caption{Different types of object detection settings to reduce the cost of data annotation}
% \label{fig.examples}
% \end{figure*}

Inspired by DETR~\cite{carion2020end} which achieves competitive performance with a single-level feature map, we propose a novel detector, Point DETR, by adding a point encoder to DETR in this paper. It can predict object boxes precisely from point annotations. Specifically, it uses a single-level feature map to predict object boxes, avoiding the multi-level selection problem and can predict object boxes with loose points, having no strict requirement on point annotations. Besides, it inherits the strong representation of DETR, having a good performance on object detection. But, different from DETR, we encode position and category of annotated points into object queries with the point encoder, which easily establishes one-to-one correspondences between points and object queries, being fit for box prediction based on points. In addition, to boost detection performance and make optimization easier, we do box predictions as offsets w.r.t. point position rather than make box predictions directly like DETR. 

To show the superiority of our detector, we mainly evaluate our proposed detector on the MS-COCO dataset~\cite{lin2014microsoft}. To make a fair comparison, we take FCOS~\cite{tian2019fcos} as the default baseline, which is regarded as a point-based detector. Following our proposed weakly semi-supervised object detection setting, object instances of small image data fraction ($5\% \sim 50\%$) are annotated fully and the rest are annotated by points. In these various settings with a different fraction of fully-annotated image data, our proposed detector outperforms other modern detectors, including multi-level feature detectors and single-level feature detectors. In particular, when using $20\%$ fully labeled data from COCO, our detector outperforms FCOS and Faster R-CNN by $2.0$ AP and $1.9$ AP, respectively.

% In our experiments, we find that our detector outperforms the baseline over $1\%$ mAP in the weakly semi-supervised detection task. Due to the fact that FCOS has a better performance than our detector only trained with supervised data, we conclude that {\bf the improvement in weakly semi-supervised detection task comes from that our proposed detector solve the problems suffered by existing detectors rather than its strong representation} (details in Section 4.x).

% which is comparable to the model trained under 60\% fully supervised data, 9.0 absolute mAP improvements against supervised counterpart, 1.9 absolute mAP improvements against a strong baseline.

\begin{figure*}[t]
\centering
\includegraphics[width=.99  \textwidth]{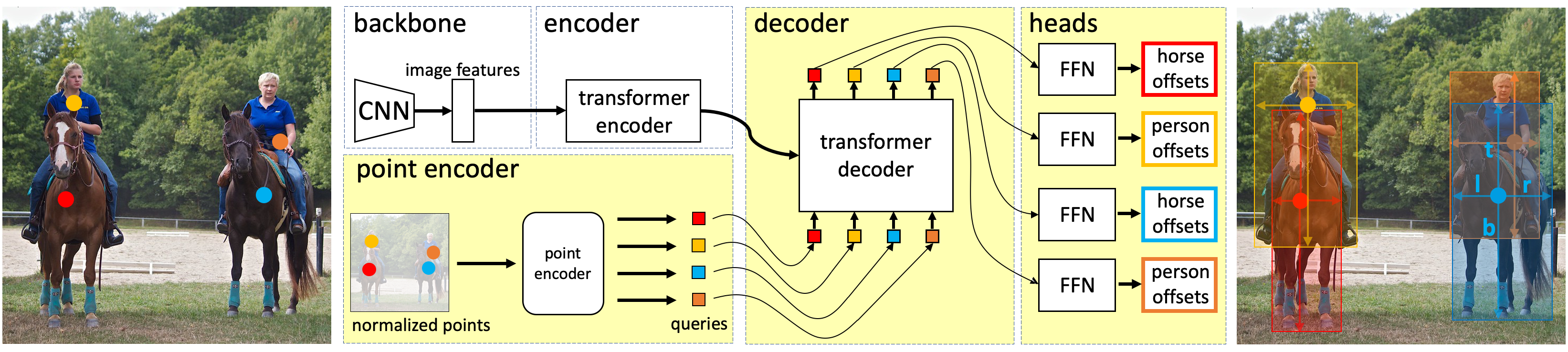}
\caption{Point DETR takes the image and its corresponding object points as input. The object points are normalized to $[0,1]^2$, and are encoded into object queries by the point encoder module. The transformer decoder takes the object queries and additionally attends to the image features (extracting by backbone and encoder). The output of the transformer decoder is passed to the head, generating box predictions. The box predictions are the relative offsets from the four sides of a bounding box to the point location. The components that are different from DETR are highlighted by light yellow.}
%The coordinates of the points are first normalized to $[0,1]$, and encoded as queries by the point encoder module, and input into the transformer decoder with the image features (extracting by backbone and transformer encoder). }
\label{fig.pipeline}\vspace{-0.1cm}
\end{figure*}

Our main contributions can be summarized as follows:
\begin{itemize}
    \item We propose a potential and novel setting for the weakly semi-supervised object detection task, which comprises small fully annotated images and large weakly annotated images by points. Compared with the image-level data setting ~\cite{bearman2016s}, this setting introduces weakly instance-level information with marginal annotation cost, which is fit for object detection. This provides a new perspective to improve detection performance with weakly annotated detection images.
    \item Based on the above setting, we analyze the drawbacks of existing modern object detectors and propose Point DETR, which is simple and easily implemented. The proposed detector takes object points as input, transforms these points into object queries, and predicts object box precisely for these queries, as shown in Figure~\ref{fig.pipeline}.
    \item Extensive experiments on COCO dataset~\cite{lin2014microsoft} are conducted to demonstrate the effectiveness of our proposed detector. Our detector outperforms most modern detectors in various data settings. We also do quantity and quality experiments to show our detector solves the problems suffered by most modern detectors.
\end{itemize}
 
%-------------------------------------------------------------------------
\section{Related Work}
\paragraph{Supervised Object Detection:} With the large-scale fully annotated detection data, existing modern detectors~\cite{law2018cornernet, lin2017feature, lin2017focal, ren2015faster, tian2019fcos} have obtained great improvements in the object detection task. These detectors can be divided into two categories: two-stage detectors and one-stage detectors. FPN~\cite{lin2017feature} is a popular two-stage detector, which predicts object proposals firstly and refines these proposals finally. Unlike two-stage detectors, one-stage detectors~\cite{law2018cornernet,lin2017focal, tian2019fcos} directly outputs the classification and location of each object without refinement. Though achieving great success, these detectors are trained with a large amount of fully-annotated data, which is costly to annotate. Thus, there are many works proposed to reduce the annotation cost.

\paragraph{Semi-Supervised/Weakly Supervised Object Detection:} Semi-supervised object detection (SSOD)~\cite{jeong2019consistency,nguyen2019semi,sohn2020simple,tang2020proposal,wang2018towards} and weakly-supervised object detection (WSOD)~\cite{bilen2016weakly,jie2017deep,shi2017weakly,zhu2017soft} are introduced to reduce the large cost of data annotation. The semi-supervised object detection methods learn detectors with a small amount of box-level labeled images and large unlabeled images. Jeong  \etal~\cite{jeong2019consistency} employ consistency constraints for object detection to exploit unlabeled data. While, weakly supervised object detection methods utilize large data with weak annotations, such as image labels. Bilen \etal~\cite{bilen2016weakly} learn an object detector under image-level supervision by combining region classification and selection. Furthermore, pursuing the performance of supervised detection and keeping the low cost of annotation, weakly semi-supervised object detection methods (WSSOD)~\cite{yan2017weakly} are studied, which use small box-level labeled images as well as large weakly labeled images to learn detectors. Unlike these semi-/weakly-supervised object detection, our proposed detector utilizes a new low-cost annotation: points, which provide instance location. Recently, UFO$^2$~\cite{ren2020ufo} also uses point supervision as weak labels, but it does not explore the point information sufficiently as we shown in Section~\ref{src.ablations.vsufo}.

\paragraph{Point based Semi-Supervised Segmentation:} Point supervision~\cite{bearman2016s, qian2019weakly, zhang2020interactive} has been employed by semantic segmentation. Bearman \etal~\cite{bearman2016s} incorporate point supervision along with objectness prior to boost segmentation performance and alleviate annotation burden. Qian \etal~\cite{qian2019weakly} leverage semantic relationships among several labeled points to address the semantic scene parsing task. Different from these works, we focus on object detection task, where point-based detection has been explored little. Due to a lack of exploitation, existing detectors do not fit point-level annotation well.

\paragraph{DETR:} Unlike existing detectors, DETR~\cite{carion2020end} removes the need for many hand-designed components like a non-maximum suppression procedure or anchor generation. By virtue of Transformer~\cite{vaswani2017attention}, DETR takes an image as input and directly outputs a fixed set of box predictions. For the point-based detection task, DETR has a beneficial characteristic: a single-level feature map, avoiding the multi-level selection problem. However, directly applied DETR into point-based detection task is not practical. Object queries in DETR are general embeddings and have no specific point information. Conversely, our detector encodes the position and category of annotated points into object queries with the point encoder and establishes one-to-one correspondences between point annotations and object queries.

\begin{figure}[t]
\centering
\includegraphics[width=0.47  \textwidth]{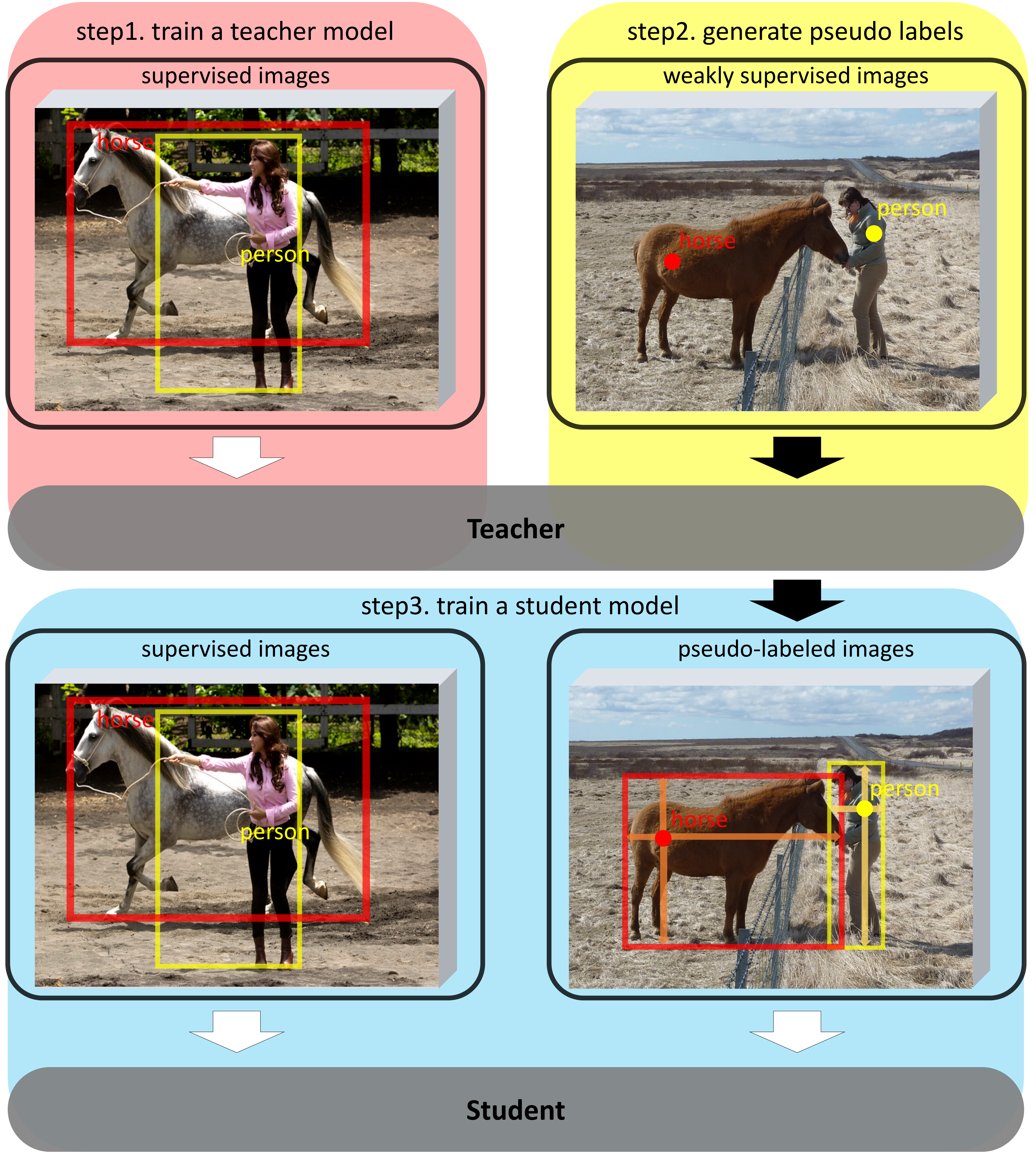}
\caption{Overall framework. The white arrows represent the training stage, and the black arrows represent the inference stage. The steps of the framework are represented by red, yellow, blue rounded rectangles respectively. Best viewed in color.}
\label{fig.framework}
\vspace{-0.2cm}
\end{figure}
% \vspace{-0.5cm}

% \begin{table}
% \begin{center}
% \begin{tabular}{|l|c|}
% \hline
% Method & Frobnability \\
% \hline\hline
% Theirs & Frumpy \\
% Yours & Frobbly \\
% Ours & Makes one's heart Frob\\
% \hline
% \end{tabular}
% \end{center}
% \caption{Results.   Ours is better.}
% \end{table}

%-------------------------------------------------------------------------
\section{Method}
% In this section, we first introduce the task of weakly semi-supervised object detection (WSSOD) with point annotations and then discuss the reasons why existing detectors can not fit the task well, whether it is a multi-level feature detector (\eg FCOS~\cite{tian2019fcos}) or 
% a single-level feature detector(\eg Faster R-CNN~\cite{ren2015faster}). Next, we review DETR~\cite{carion2020end} and explain why it is not suitable for the task. Finally, we illustrate our method Point DETR in detail.

In this section, we first introduce the task of weakly semi-supervised object detection (WSSOD) with point annotations and discuss why existing object detectors can not fit this task well. Next, in order to solve it, we illustrate our novel detector, Point DETR, in detail.

%the overall framework of our approach, and then review DETR~\cite{carion2020end} in details as it is highly related to our method, and explain why it is not suitable for our task. In section 3.3, we propose our point-level detector and highlight several components in this detector.
\paragraph{WSSOD with point annotations:} WSSOD generally uses a small set of instance-level labeled images and tremendous weakly image-level labeled images as training data (Figure~\ref{fig.examples}c). However, for object detection, image-level labeled images do not fit WSSOD well, since it can not provide instance information. This raises a natural question: is there a new data annotation for weakly labeled images, which has instance information without a large annotation burden? In this paper, we introduce point annotation for weakly labeled images.
% \vspace{-0.1cm}
\paragraph{Point Annotations:} It is introduced by Bearman \etal~\cite{bearman2016s} for weakly semantic segmentation, but it has not been explored well in object detection. In object detection, we define point annotation as follows: it locates on the object and takes object class as its category. Thus, we represent an object as $(x,y,c)$, where $(x, y) \in [0,1]^2$ and $c$ represent point location and object category, respectively. We must note that our method is robust to point location, as shown in Table~\ref{tab.label.position}. Therefore, the point annotations can locate at the anywhere of objects. In this way, we can alleviate the annotation burden.
% Even if we do not limit the position of the labeling point (~\cite{papadopoulos2017training,roy2018deep,ren2020ufo} \etc)
% Unlike ~\cite{papadopoulos2017training,roy2018deep,ren2020ufo} and other tasks that require annotator to label the center point of an imaginary bounding box, our point annotations have less constraint under our definition.  
% In this way, we can alleviate the annotation burden and make our method {\color{red}robust} to point location. 
% In section 4.x, we conduct experiments on labeling quality, which shows that with the improvement of labeling quality, the effect of our method can be further increased.

\paragraph{Overall Framework:} With this new setting that a small number of supervised images and a large number of weakly supervised images,
% that a small set of labeled images and large weakly point labeled images, 
we adapt self-training as our default training pipeline, which has made considerable progress in semi-supervised learning (\eg Lee~\cite{lee2013pseudo},Noise-Student~\cite{xie2020self},STAC~\cite{sohn2020simple}). The steps are summarized as follows:

\begin{enumerate}
  \item {\bf Train a teacher model} on available labeled images. 
  \item {\bf Generate pseudo-labels of weakly point annotated images} using the trained teacher model. 
  \item {\bf Train a student model} with fully labeled images and pseudo-labeled images.
\end{enumerate}

%We focus on this step in our paper and propose a point-level detector that predicts precise bounding box from points (details in section 3.3). For pseudo-label based methods (\eg,~\cite{sohn2020simple,lee2013pseudo,yan2017weakly,xie2020self}), a hyperparameter carefully selected manually is required to screen out pseudo-labels with sufficiently high confidence while taking the recall into account. By taking advantage of the strong prior knowledge about the existence of objects brought by points, the hyperparameter is not needed in our method, as we referred to in section 4.x.

%We {\bf do not} look the pseudo-labeled images differently, so no need for extra hyperparameter to control the ratio between supervised loss and weakly supervised loss, unlike CSD~\cite{jeong2019consistency} or STAC~\cite{sohn2020simple}. Further, We regard the unified student model's performance as our evaluation metric, therefore we only introduce weak augmentation (\ie horizontal flip and the images are resized to a maximum scale of $1333 \times 800$ without changing the aspect ratio, following common practices~\cite{chen2019mmdetection,tian2019fcos,lin2017feature}) to train the student model. 这段放到experiments

The overall framework is shown in Figure~\ref{fig.framework}. For most self-training based detection methods, hyper-parameters are selected carefully since they must keep true object boxes and screen out false ones as much as possible. Instead, we can directly predict the corresponding object box for each point annotation without duplicate object boxes. Although choosing hyper-parameters is no longer an obstacle to performance, predicting object boxes from point-level annotations with existing detectors remains a problem.

\paragraph{Discussion on Existing Detectors:} Existing detectors can be divided into two categories: multi-level feature detectors and single-level feature detectors. For multi-level detectors (\eg FCOS\cite{tian2019fcos}), it is difficult for them to predict object boxes with point annotations since point annotations do not have feature-level information, which is used to select one prediction from multi-level box predictions (Figure~\ref{fig.vs.world}b). On the other hand, single-level feature detectors (\eg Faster R-CNN\cite{ren2015faster}) suffer from the bad performance or strict requirement on point annotations though avoiding choosing feature map levels (Figure~\ref{fig.vs.world}c). For more experiments see Section~\ref{src.ablations.singlelevel}.
%This is very similar to Figure~\ref{fig.examples.SSOD-example}) and Figure~\ref{fig.examples.WSSOD-example} where a set of fully supervised images are supported. A nature idea is to learn from methods that have achieved success in these settings. Fortunately, Lee~\cite{lee2013pseudo} proposed a simple and efficient method: Pseudo-label that worked well both in those settings[xxx, xxx, xxx]. Through inspiration from STAC~\cite{sohn2020simple}, we extended its framework to our setting based on pseudo-label, avoiding introducing extra hyperparameters. The steps and the main differences from STAC are summarized as follows:

\subsection{Point DETR}
To avoid the drawbacks of existing detectors in the WSSOD with point annotations task, we introduce a novel detector, Point DETR: adding a point encoder to DETR. It transforms point annotations into object queries, extracts image features for each object query, and outputs the corresponding object box. Next, we introduce a key element of Point DETR, point encoder, which is critical to the WSSOD with point annotations task.

\paragraph{DETR:} We begin by reviewing DETR~\cite{carion2020end}, which is an end-to-end set-based object detector. DETR consists of a CNN backbone, an encoder-decoder transformer, and a prediction head. DETR first extracts a single-level 2D feature map from the CNN backbone, flattens it, and supplements it with a positional encoding. Then, the encoder-decoder transformer takes as input a fixed set of object queries (learned positional embeddings) and attends to 1D image feature embeddings. Finally, the output embeddings of the transformer are passed to the prediction head that predicts either a detection (class and bounding box) or a  ``no object'' class.

\paragraph{Point DETR:} Point DETR, as shown in Figure~\ref{fig.pipeline}, adopts most components of DETR. To fit the point annotated images, Point DETR has a special module, point encoder. Point encoder can encode the point annotations into object queries, which are taken as input by the transformer decoder. Unlike the object queries in DETR that are learned positional embeddings, these object queries are specific instance embeddings which contain position and category information of object instances. Thus, these object queries have a one-to-one correspondence with object instances. Moreover, the number of object queries varies with the number of object instances in an image instead of a fixed number(e.g. 100) like DETR.

During training, we simply define the loss of each object query as $\mathcal{L}=\mathcal{L}_{box}$, since we already have category for each object query and only need to regress the object box. The bounding-box loss $\mathcal{L}_{box}$ is identical as it defined in DETR. But, for the box prediction $\hat{b}_{i}$, it calculated by $\hat{b}_{i}=\hat{b}_{i}^{init}+\Delta \hat{b}_{i}$, where $\hat{b}_{i}^{init} \in [0,1]^4$ is $(x, y, x, y)$, $(x, y)$ is the location of point annotation and $\Delta \hat{b}_{i} \in [0,1]^4$ is the relative offsets \wrt the point location $(x, y)$ following FCOS~\cite{tian2019fcos}. In our experiments, we show this way of regression can alleviate the mismatch between point annotation and object box, see Section~\ref{sec.ablations.regress}.

\begin{figure}[t]
\centering
\includegraphics[width=0.47\textwidth]{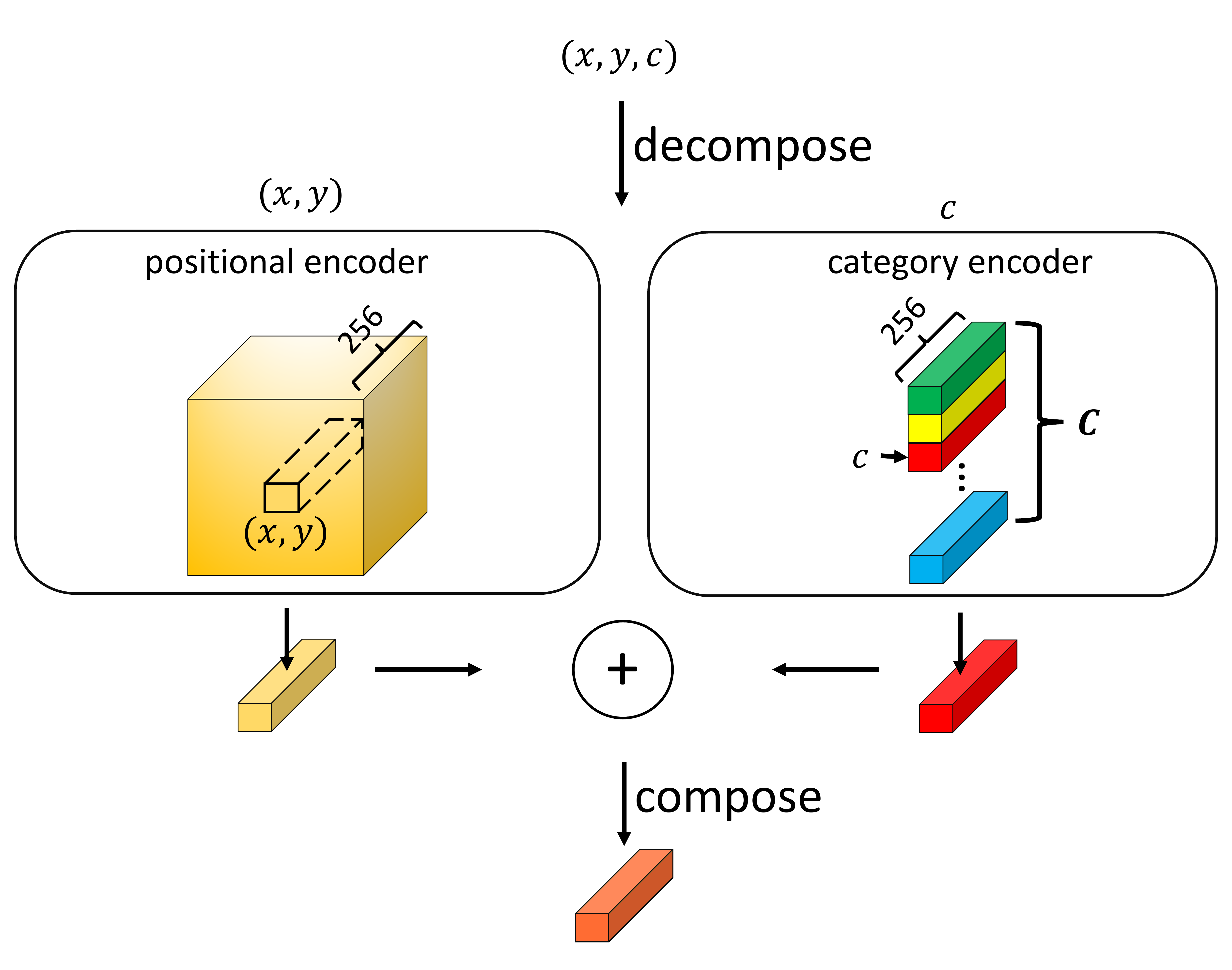}
\caption{Point encoder. For each point $(x,y,c)$, it encodes the position $(x,y)$ and category $c$ separetely, and then takes the element-wise addition as the point embedding.}
\label{fig.query-encoder}
\vspace{-0.1cm}
\end{figure}

\paragraph{Point Encoder:} In point DETR, how to encode point annotations into object queries is critical for point encoder. As shown in Figure~\ref{fig.query-encoder}, a point annotation $(x, y, c)$ is decomposed to a 2D coordinate $(x,y) \in [0,1]^{2}$ and category index $c$. Based on $(x,y)$, the position embedding $e_{pos} \in \mathbb{R}^{256}$ is extracted from fixed spatial positional encodings~\cite{vaswani2017attention,parmar2018image,carion2020end}, which is the same as one used in the transformer encoder. For category embedding $e_{cat} \in \mathbb{R}^{256}$, it is obtained from predefined learnable category embeddings by category index, \ie $c$. In the end, we fuse these embedding to get the object query by sum operation. 

Though point encoder is simple and easily implemented,  it bridges the divisions between point annotations and object queries. In the experiments, we show the essentials of every component (positional encoder and category encoder) in point encoder, see section~\ref{sec.ablations.point.encoder}.

% A point, noted by $(x,y,c)$, could be decomposed to a 2D coordinate $(x,y)$ and category index $c$. In our work, we encode them separately, and then add the encoded position and category element wisely, as we have shown in Figure~\ref{fig.query-encoder}. We use fixed spatial positional encodings~\cite{vaswani2017attention,parmar2018image,carion2020end} which is also applied in the transformer module, to maintain the consistency of the spatial position embedding in the entire detector. For category encoding, due to the limited number of categories, we have predefined learnable embeddings for all categories and obtain the embedding by category index, \ie $c$.

\begin{figure}[t]
\centering
\includegraphics[width=0.47\textwidth]{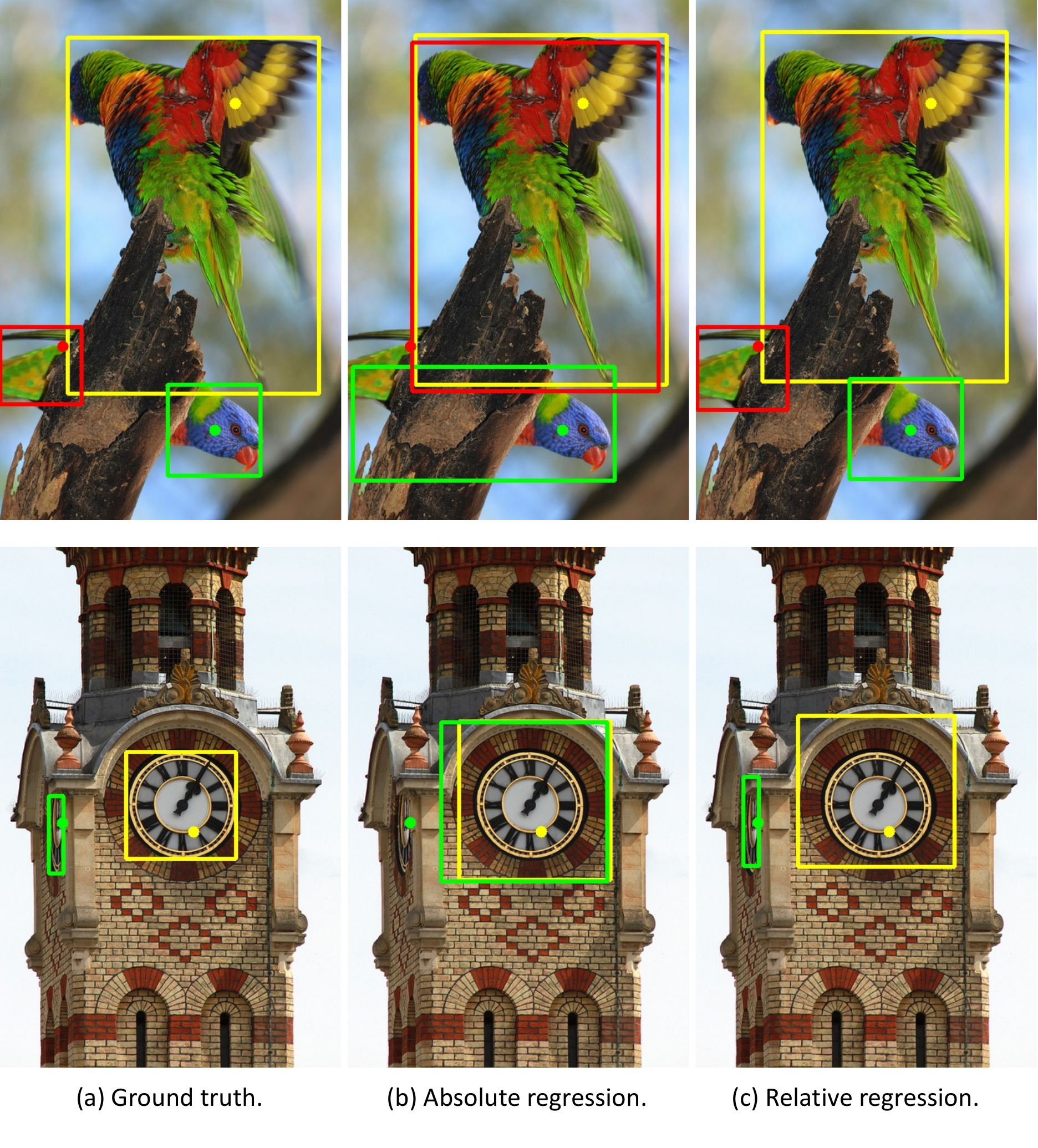}
% \begin{subfigure}[t]{0.325\linewidth}
% \includegraphics[width=1.\textwidth]{figures/gt.jpg}
% \includegraphics[width=1.\textwidth]{figures/gt_clock.pdf}
% % \captionsetup{font=small}
% \setcaptionwidth{.95\textwidth}
% \captionsetup{font=footnotesize}
% \caption{Ground truth.}
% \end{subfigure}
% \begin{subfigure}[t]{0.325\linewidth}
% \includegraphics[width=1.\textwidth]{figures/xywh.jpg}
% \includegraphics[width=1.\textwidth]{figures/xywh_clock.pdf}
% % \captionsetup{font=small}
% \setcaptionwidth{.95\textwidth}
% \captionsetup{font=footnotesize}
% \caption{Absolute regression.}
% \label{fig.xywh}
% \end{subfigure}
% \begin{subfigure}[t]{0.325\linewidth}
% \includegraphics[width=1.\textwidth]{figures/our.jpg}
% \includegraphics[width=1.\textwidth]{figures/our_clock.pdf}
% % \captionsetup{font=small}
% \setcaptionwidth{.95\textwidth}
% \captionsetup{font=footnotesize}
% \caption{Relative regression.}
% \label{fig.ltrb}
% \end{subfigure}
\caption{\textbf{Absolute vs. Relative Regression:} Different colors to distinguish instances and the color of the point annotation is consistent with its corresponding box. Best viewed in color.}
\label{fig.ltrb.vs.xywh}\vspace{-0.5cm}
\end{figure}

%{\bf Boxes to points}. In the training stage, our task is transformed into using fully labeled data (\ie annotated by a bounding box) to train the point-level detector. We simply sample points from the ground truth boxes according to a uniform distribution.  这段放到experiments里面去。

%-------------------------------------------------------------------------

\section{Experiments}
We evaluate our models on the COCO 2017 detection dataset~\cite{lin2014microsoft} with synthetic point annotations (details in section 4.1). We report the standard COCO metrics including $AP$ (averaged over IoU thresholds), $AP_{50}$, $AP_{75}$. In addition, to show the quality of generated pseudo-boxes, we also calculate the $mIoU$ between the generated pseudo-boxes and the ground truth bounding boxes.

With existing detectors that can not be directly applied to our point annotated settings, we make some modifications to existing detectors: FCOS and Faster R-CNN. These modified detectors are denoted as FCOS$^\dagger$ and Faster R-CNN$^\dagger$, respectively. For FCOS{$^\dagger$}, we separately extract point features from multi-level feature maps by bilinear interpolation~\cite{jaderberg2015spatial} and predict the corresponding object box, finally use the box prediction with the highest point category score as the pseudo-box. As for Faster R-CNN{$^\dagger$}, we extract point features from one-level feature map, and then predict boxes for different anchors, finally use one with the highest point category score as the pseudo-box.

\subsection{Implementation Details}
We use ResNet-50~\cite{he2016deep} as the default backbone for different detectors and set the hyper-parameters following these detectors.

% \begin{table*}
% \begin{center}
% \begin{tabular}{l|| l |l | l| l| l| l}
% % \hline
%  COCO & 5\% & 10\% & 20\% & 30\% & 40\%  & 50\%  \\
% \hline
% Baseline & 17.2 & 21.1 & 25.0 & 28.5 & 30.6 & 32.8 \\
% \hline
% FCOS & 25.5 & 28.4 & 31.3 & 32.8 & 33.8 & 34.5 \\
% % \hline 
% Point DETR ({\bf ours}) & 26.2 ({\bf+0.7}) & 30.4 ({\bf+2.0}) & 33.3 ({\bf+2.0}) & 34.8 ({\bf+2.0}) & 35.4 ({\bf+1.6}) & 35.8 ({\bf+1.3}) \\
% % \hline
% \end{tabular}
% \end{center}
% \caption{Comparison in mAPs of the student model (\ie FCOS) for different methods on MS-COCO. ``Supervised" refers to the student models trained on labeled data only.}
% \label{tab.ours.vs.baseline}
% \end{table*}

% \begin{table*}
% \begin{center}
% \begin{tabular}{l|| l |l | l| l| l| l}
% % \hline
%  COCO & 5\%  & 10\%  & 20\% & 30\% & 40\%  & 50\%  \\
% \hline
% FCOS$^\dagger$~\cite{tian2019fcos}& 20.9 & 25.4  &30.9 & 33.3  &34.9  &35.4 \\
% % \hline
% DETR~\cite{carion2020end}& 18.2 ({\bf-2.7}) & 23.7 ({\bf-1.7}) & 29.3 ({\bf-1.6}) & 32.9 ({\bf-0.4}) & 34.8 ({\bf-0.1}) & 36.3 ({+0.9}) \\
% % \hline
% \end{tabular}
% \end{center}
% \caption{Comparison in mAPs of FCOS$^\dagger$ and DETR to demonstrate the improvement comes from our method rather than a stronger teacher model. ``FCOS$^\dagger$" indicates the FCOS model trained in DETR augmentation. In most cases (5\%-40\%), FCOS$^\dagger$ has a better performance than DETR.}\label{tab.fcos.vs.detr}
% \end{table*}

\paragraph{Dataset:}
We train the model with 118k training images and evaluate the performance of the detectors on the remaining 5k val images. Specially, for our point annotated setting, we randomly sample 5\%, 10\%, 20\%, 30\%, 40\%, 50\% of training images as the fully labeled set and use the rest of the images as a weakly labeled set. In this paper, we noted them as different data settings for simplicity, \eg 20\% data setting. For the weakly labeled set, we synthesize the point annotations for each object as follows: (a) if the object has instance segmentation, randomly sample a point from the instance mask as the point annotation for the object; (b) if not, simply randomly sample a point in its bounding box.

\paragraph{Training:}
In our framework, there are two models: the teacher model and student model. Our teacher model includes Point DETR, FCOS{$^\dagger$}, and Faster R-CNN{$^\dagger$}. While we simply choose FCOS as the default student model since the student model is only used to evaluate the effectiveness of the teacher model. We show by experiments (in section~\ref{sec.ablations.student}) that our method is robust to the architecture of the student model.

For the training of the teacher model, it is simple for FCOS{$^\dagger$} and Faster R-CNN{$^\dagger$}. We train them with their default training settings. For a fair comparison, we also use data augmentation as shown in ~\cite{carion2020end}. For Point DETR, it follows most of the training settings used in~\cite{carion2020end} with several differences: we train the model for 108 epochs on 8 GTX 1080Ti GPUs, with 2 images per GPU. To ensure training stability, we use a warmup scheme~\cite{goyal2017accurate} in the first epoch. The learning rate is reduced by a factor of 10 at epoch 72 and 96, respectively. In the training, we randomly sample a point in each bounding box and transform points into point annotations. With these point annotations, we train Point DETR as shown in Figure~\ref{fig.pipeline}. 

For the default student model, we combine the fully labeled images and pseudo-labeled images generated by the teacher model to train the student, as showed in Figure~\ref{fig.framework}. 

\begin{figure}[t]
\centering
\includegraphics[width=.47\textwidth]{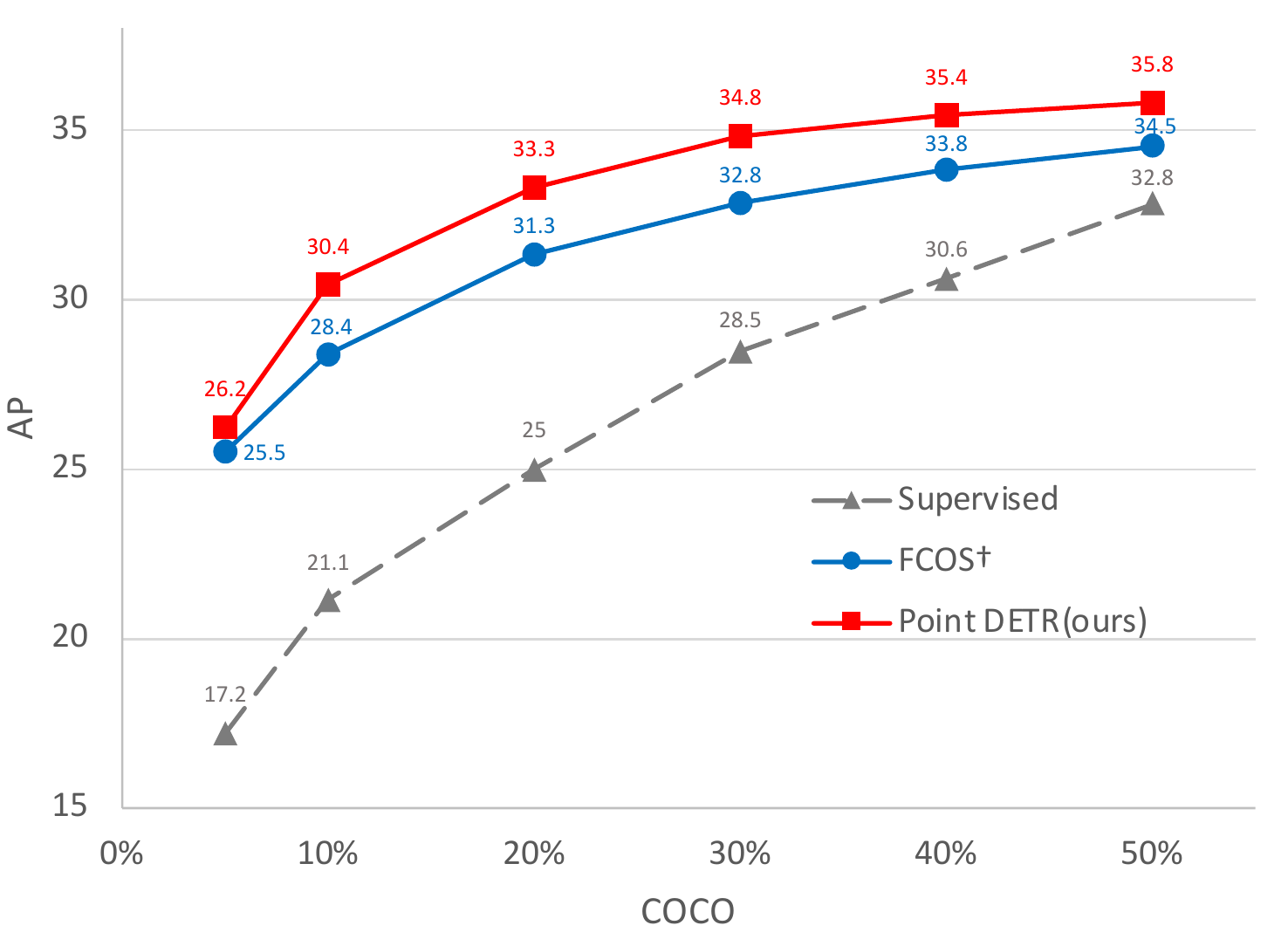}
\captionsetup{font=small}
% \setcaptionwidth{.9\textwidth}
% \vspace{-0.5cm}
\caption{Comparison in APs of the student model (\ie FCOS) for different methods on MS-COCO. ``Supervised" refers to the student models trained on labeled data only.}\label{fig.results.ours.vs.baseline}
\vspace{-0.30cm}
\end{figure}

\subsection{Main Results}\label{section.main.results}
We first show the effectiveness of Point DETR on different data split settings, see Figure~\ref{fig.results.ours.vs.baseline}. We train the student model (\ie FCOS) only with the fully annotated images (noted as ``Supervised''). By comparing ``Supervised'' with the student model trained with pseudo-boxes, we can evaluate the benefits brought by the pseudo-boxes. Point DETR and FCOS$^\dagger$ outperform ``Supervised'' by a large margin. This demonstrates that images with point annotations can improve the performance of the detection task. Furthermore, Point DETR outperforms FCOS$^\dagger$ by a considerable margin. 

\begin{figure}[t]
\centering
\includegraphics[width=.47\textwidth]{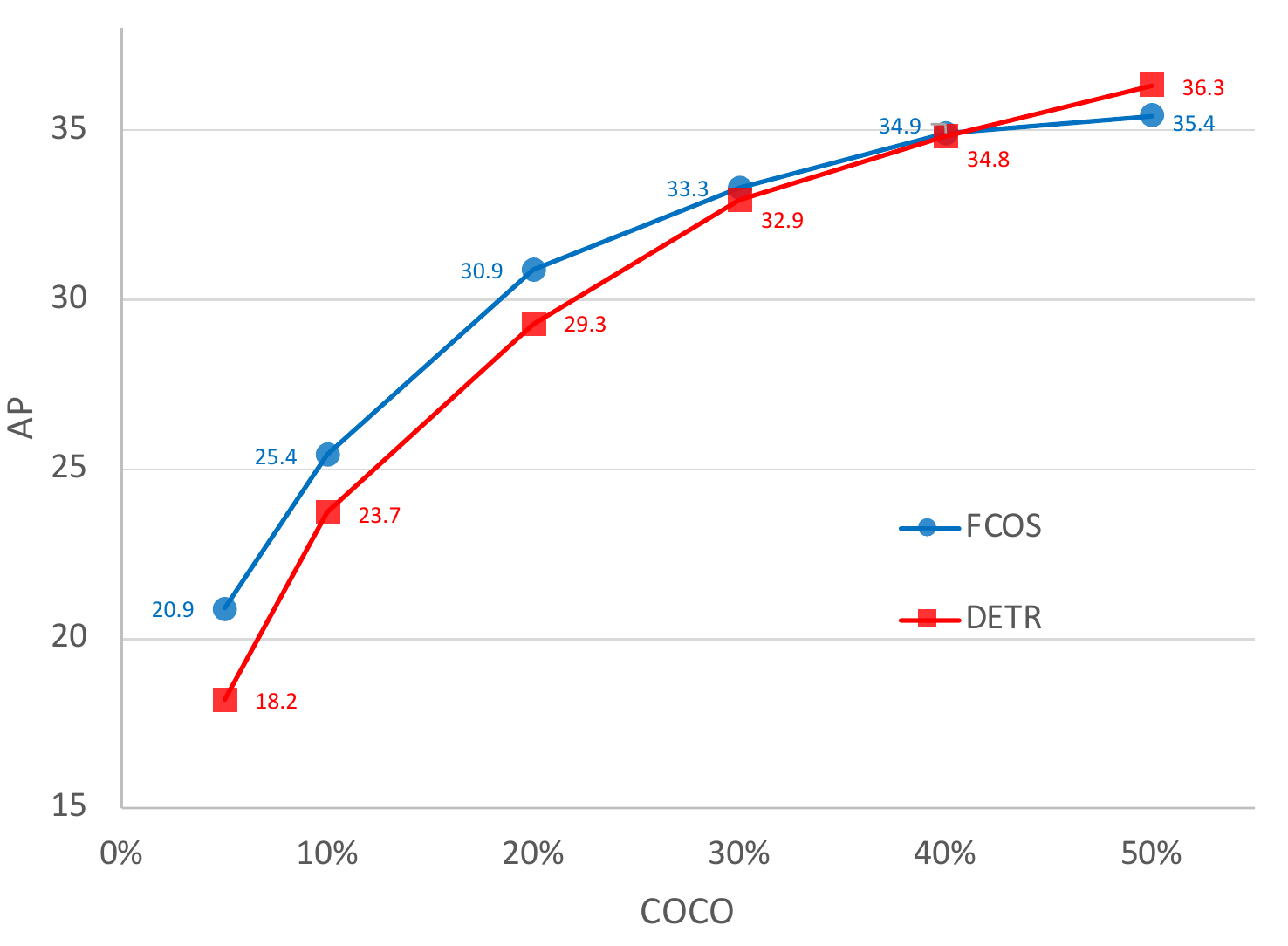}
\captionsetup{font=small}
% \setcaptionwidth{.4\textwidth}
% \vspace{-0.5cm}
\caption{Comparison in APs of FCOS and DETR to demonstrate the improvement comes from our method rather than a stronger teacher model. FCOS trained in DETR augmentation for a fair comparison. In most cases ($5\% \sim 40\%$), FCOS has a better performance than DETR.}
\label{fig.results.detr.vs.fcos}
% \vspace{-0.30cm}
\end{figure}

\begin{table*}[t]\vspace{0mm}
\footnotesize
\centering
\subfloat[\textbf{Point Encoder:} The effectiveness of positional encoder and category encoder.\label{tab.ablations.pointencoder}]{
\begin{minipage}{.37\textwidth}
\centering
\tablestyle{5pt}{1.05}\setlength{\tabcolsep}{1.mm}\begin{tabular}{c|c|c|ccc}
 \scriptsize  & pos? & cate? & AP & AP$_{50}$ & AP$_{75}$\\
\shline
 \scriptsize \multirow{3}*{Point Encoder} &&\checkmark& 14.7&34.3&10.4 \\
 \scriptsize ~ &\checkmark&  & 31.3&51.0&32.6\\
 \scriptsize ~ &\checkmark&\checkmark& \textbf{33.3}&\textbf{53.5}&\textbf{34.8}\\
\end{tabular}
\end{minipage}
}\hspace{1cm}
\subfloat[\textbf{Student Model:} RetinaNet~\cite{lin2017focal} as the student model demonstrates the effectiveness of our approach is not related to the student model.\label{tab.ablations.retina.student}]{
\begin{minipage}{.45\textwidth}
\centering
\tablestyle{5pt}{1.05}\setlength{\tabcolsep}{1.mm}\begin{tabular}{cc||ccc}
 \scriptsize  Teacher & Student&AP&AP$_{50}$&AP$_{75}$\\
\shline
 \scriptsize FCOS$^{\dagger}$ &\multirow{2}*{RetinaNet}& 30.4& 49.9 & 31.6\\
 \scriptsize Ours &~& \textbf{32.5} & \textbf{52.8} & \textbf{33.7}\\
 \hline
 \scriptsize FCOS$^{\dagger}$&\multirow{2}*{FCOS} & 31.3 & 50.7 & 32.6\\
 Ours&~&\textbf{33.3}&\textbf{53.5}&\textbf{34.8}\\
\end{tabular}
\end{minipage}
}

\subfloat[\textbf{Single-Level Detector:} Point DETR vs. Faster R-CNN$^\dagger$.\label{tab.ablations.fasterrcnn}]{
\begin{minipage}{.25\textwidth}
\centering
\tablestyle{5pt}{1.05}\setlength{\tabcolsep}{1.mm}\begin{tabular}{c||ccc}
 \scriptsize  & AP & AP$_{50}$ & AP$_{75}$ \\
\shline
 \scriptsize  Faster R-CNN$^\dagger$& 31.4&51.6&32.6\\
 \scriptsize Ours & \textbf{33.3} &\textbf{53.5}&\textbf{34.8} \\
\end{tabular}
\end{minipage}
}\hspace{.66cm}
\subfloat[\textbf{Comparison with UFO$^2$~\cite{ren2020ufo}:} ``Supervised'' refers to the model trained with fully labeled data only.\label{tab.ours.vs.ufo}]{
\begin{minipage}{.3\textwidth}
\tablestyle{5pt}{1.05}\setlength{\tabcolsep}{1.mm}\begin{tabular}{c||c| c c c}
 \scriptsize  & Supervised&AP & AP$_{50}$ & AP$_{75}$ \\
\shline
 \scriptsize  UFO$^2$  &29.1& 30.1 & - & -\\
 \scriptsize Ours  & 28.1&\textbf{33.5} & 53.8 & 34.8\\
\end{tabular}
\end{minipage}
}\hspace{.66cm}
\subfloat[\textbf{Point Location:} The effectiveness of the point location.\label{tab.label.position}]{
\begin{minipage}{.25\textwidth}
\tablestyle{5pt}{1.05}\setlength{\tabcolsep}{1.mm}\begin{tabular}{c||c| c c c}
 \scriptsize  &center?&AP & AP$_{50}$ & AP$_{75}$ \\
\shline
 \scriptsize Ours & &\textbf{33.3} & 53.5 & \textbf{34.8}\\
 \scriptsize Ours & \checkmark &33.3 & \textbf{53.6} & 34.6\\
\end{tabular}
\end{minipage}
}

\centering
\subfloat[\textbf{Point Annotations:} To confirm the benefits of point annotations, we compare Point DETR (with points) vs. DETR (without points) by analyzing the generated boxes with respect to ground truth boxes. With AR far exceeding DETR, our AP remains comparable.\label{tab.ablations.pseudo.quality}]{
\tablestyle{5pt}{1.05}\begin{tabular}{c||c|c|ccc|ccc|ccc|ccc}
 \scriptsize  & points? &score? &AP&AP$_{50}$&AP$_{75}$&AP$_{s}$&AP$_{m}$&AP$_{l}$&AR$_{1}$&AR$_{10}$&AR$_{100}$&AR$_{s}$&AR$_{m}$&AR$_{l}$\\
\shline
 \scriptsize \multirow{2}*{DETR}& & & 19.1 &33.2 &18.7& 5.6&20.2&31.3&22.9&32.8&33.6&12.0&35.0&51.0\\
 \scriptsize ~& &\checkmark&\textbf{26.9}&43.8&\textbf{27.5}&9.2&27.7&\textbf{39.9}&24.2&33.2&33.6&12.0&35.0&51.0\\
 \hline
 \scriptsize Ours&\checkmark&&26.8&\textbf{52.3}&24.2&\textbf{12.6}&\textbf{29.5}&38.9&\textbf{30.7}&\textbf{44.0}&\textbf{44.5}&\textbf{22.8}&\textbf{46.9}&\textbf{63.9}\\
 \hline
 $\Delta$&&&-0.1&+8.5&-3.3&+3.4&+1.8&-1.0&+6.5&+10.8&+10.9&+10.8&+11.9&+12.9\\
\end{tabular}}\hspace{3mm}
\vspace{0.2cm}
\caption{\textbf{Ablations.} All ablation experiments are conducted at 20\% data setting except (d).}\label{tab.ablations.total}
\end{table*}

\begin{figure*}[t]
\centering
\includegraphics[width=.97\textwidth]{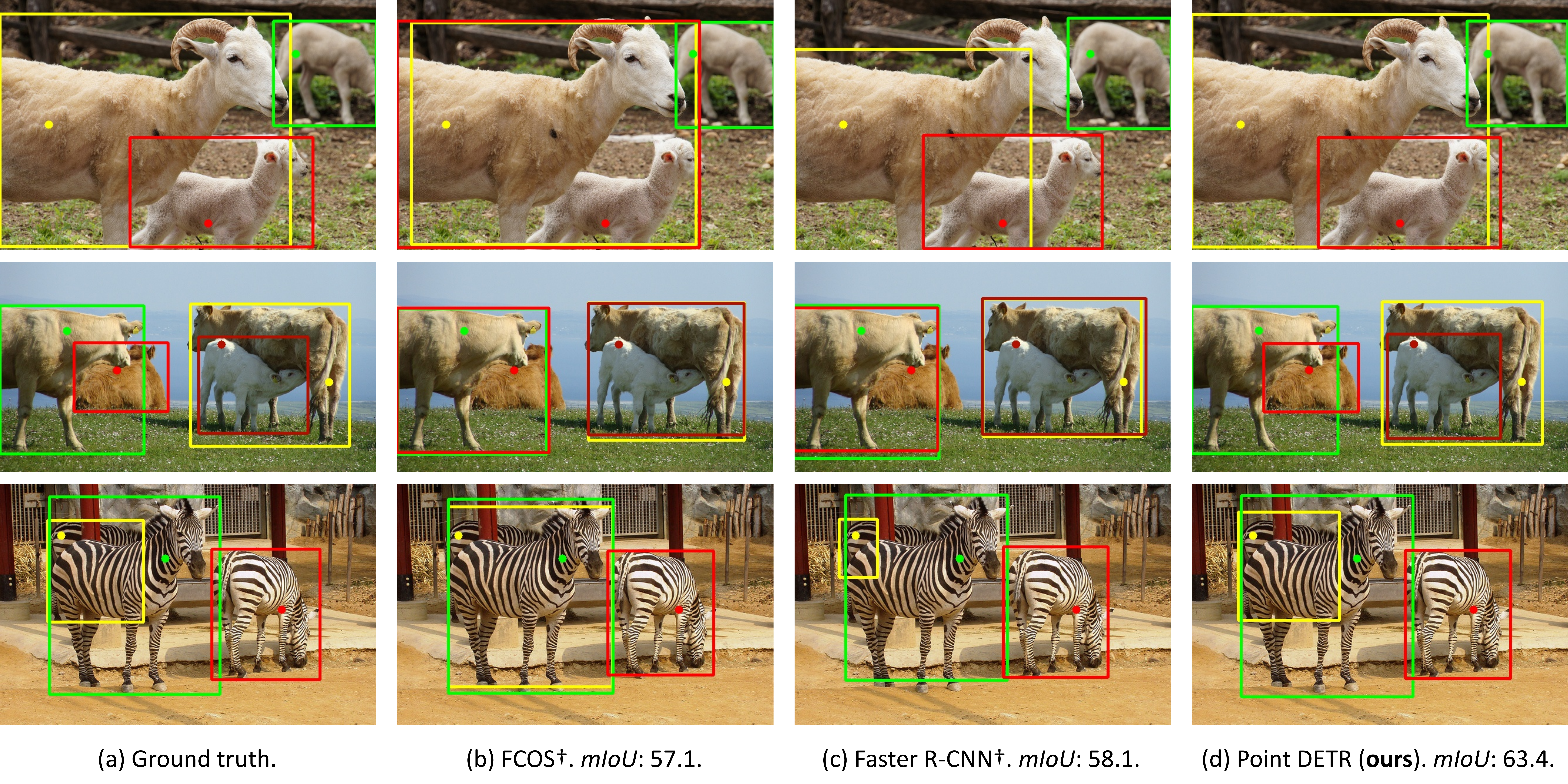}
% \begin{subfigure}[t]{0.245\linewidth}
% \includegraphics[width=1.\textwidth]{figures/goat/gt.jpg}
% \includegraphics[width=1.\textwidth]{figures/milk/gt.pdf}
% \includegraphics[width=1.\textwidth]{figures/zebra/gt.jpg}
% \captionsetup{font=small}
% % \setcaptionwidth{.9\textwidth}
% \caption{Ground truth.}
% \end{subfigure}
% \begin{subfigure}[t]{0.245\linewidth}
% \includegraphics[width=1.\textwidth]{figures/goat/fcos.jpg}
% \includegraphics[width=1.\textwidth]{figures/milk/fcos.pdf}
% \includegraphics[width=1.\textwidth]{figures/zebra/fcos.jpg}
% \captionsetup{font=small}
% % \setcaptionwidth{.9\textwidth}
% \caption{FCOS$^{\dagger}$. $mIoU$: 57.1.}\label{fig.vs.world.fcos}
% \end{subfigure}
% \begin{subfigure}[t]{0.245\linewidth}
% \includegraphics[width=1.\textwidth]{figures/goat/fasterrcnn.jpg}
% \includegraphics[width=1.\textwidth]{figures/milk/fasterrcnn.pdf}
% \includegraphics[width=1.\textwidth]{figures/zebra/fasterrcnn.jpg}
% \captionsetup{font=small}
% % \setcaptionwidth{.9\textwidth}
% \caption{Faster R-CNN$^{\dagger}$. $mIoU$: 58.1.}\label{fig.vs.world.fasterrcnn}
% \end{subfigure}
% \begin{subfigure}[t]{0.245\linewidth}
% \includegraphics[width=1.\textwidth]{figures/goat/our.jpg}
% \includegraphics[width=1.\textwidth]{figures/milk/our.pdf}
% \includegraphics[width=1.\textwidth]{figures/zebra/our.jpg}
% \captionsetup{font=small}
% % \setcaptionwidth{.9\textwidth}
% \caption{Point DETR (\textbf{ours}). $mIoU$: 63.4.}\label{fig.vs.world.ours}
% \end{subfigure}
\captionsetup{font=small}
% \setcaptionwidth{.4\textwidth}
\caption{Visualized results of FCOS$^{\dagger}$, Faster R-CNN$^{\dagger}$ and Point DETR (~\textbf{ours}). The $mIoU$ between the ground truth boxes and pseudo-boxes on the entire weakly labeled images are provided. Different colors to distinguish instances, and the color of the point annotation is consistent with its corresponding box. Best viewed in color. }\label{fig.vs.world}
% \vspace{-0.4cm}
\end{figure*}

Next, we verify the factors that contribute to the great performance of our method. We compare the accuracies of FCOS and DETR as shown in Figure~\ref{fig.results.detr.vs.fcos}, DETR performs worse than FCOS in most settings. Given that our method based on DETR achieves greater performance, we can conclude that the high accuracy of our method does not mainly benefit from its strong representation. Moreover, we conduct quality and quantity experiments to show the superiority of our method on pseudo-object boxes, see Figure~\ref{fig.vs.world}. FCOS{$^\dagger$}, a multi-level feature detector, can not predict object boxes well due to FPN, and Faster R-CNN{$^\dagger$}, a single-level feature detector, also has difficulty regressing box owing to poor representation. But, Point DETR can generate a more precise object box than other detectors. Specifically, the $mIoU$ of Point DETR is larger than FCOS{$^\dagger$} and Faster R-CNN{$^\dagger$} by 6.3 and 5.3, respectively. Based on the above experiments, our method achieves considerable performance mainly by generating precise pseudo-object boxes from point annotations.

\subsection{Ablation Experiments}\label{sec.ablations}
We conduct the ablation experiments at 20\% data setting. Results are shown in Table~\ref{tab.ablations.total} and discussed in detail next. 

\paragraph{Point Encoder:}\label{sec.ablations.point.encoder} Table~\ref{tab.ablations.pointencoder} shows the effectiveness of the components in the point encoder module (as we shown in Figure~\ref{fig.query-encoder}) . Point DETR with only positional embeddings outperforms one with only category embeddings and point DETR has a severe loss in AP ($18.6$ points) without positional embeddings. Based on that our method only regresses the object boxes, this suggests that it is difficult to learn the relative offsets of a point with respect to the bounding box without positional embeddings. We also find that adding category embeddings to positional embeddings can boost the performance by 2 points. We conjecture this improvement is caused by the that category embeddings can provide object prior, such as object shape.
\paragraph{Student Model:}\label{sec.ablations.student} For student model, we use FCOS~\cite{tian2019fcos} as the default detector. To exploit robustness of our approach, we replace FCOS with RetinaNet~\cite{lin2017focal}. In Table~\ref{tab.ablations.retina.student}, we find that our method has a 2.1 AP gain over baseline. This demonstrates that our method is robust to the student model.
% \vspace{-0.5cm}
\paragraph{Single-Level Detector:}\label{src.ablations.singlelevel} We compare Point DETR with single-level feature detectors and choose Faster R-CNN$^\dagger$ as the default single-level feature detector. As shown in Table~\ref{tab.ablations.fasterrcnn}, Point DETR outperforms Faster R-CNN$^\dagger$ by 1.9 points. This highlights that effectiveness of Point DETR. 

\paragraph{Comparison with UFO$^2$~\cite{ren2020ufo}:}\label{src.ablations.vsufo}
To show the effectiveness of our method, we compare Point DETR with UFO$^2$. For fair comparison, we train Point DETR following the dataset split in UFO$^2$: COCO-35 (fully labeled images) and COCO-80 (point labeled images). As shown in Table~\ref{tab.ours.vs.ufo}, our method has inferior performance than UFO$^2$ when is only trained on COCO-35, but it outperforms UFO$^2$ by 3.4 points adding COCO-80. This indicates that our method can make better use of point annotation information.
% UFO$^2$ splits the COCO 2017 detection dataset to COCO-35 (with 35k fully labeled images) and COCO-80 (with 80k weakly labeled images). We train Point DETR following the training setting in UFO$^2$. Concretely, with the data augmentation same as UFO$^2$, we train a teacher model with fully labeled images in COCO-35, and with the help of weakly labeled images (labeled by points) in COCO-80, we train a student model and evaluate its performance on the val set. The results are shown in Table~\ref{tab.ours.vs.ufo}. With the help of points, the performance of Point DETR exceeds UFO$^2$ by 3.4 AP points. This gap does not come from the student model as with only fully labeld images, the student model is inferior to UFO$^2$ (28.1 vs. 29.1). This indicates that our method can make better use of point annotation information.

\paragraph{Point Location:} 
To validate that our method is robust to the point location, we compare performances between two point location schemes: center point and arbitrary point on objects. As shown in Table~\ref{tab.label.position}, our method has comparable performance between these two point location scheme.
% We generate pseudo-boxes from its center points, while the performance is almost the same regardless of the point location as we shown in Table~\ref{tab.label.position}. This indicates that our method is robust to the point location.

% \paragraph{Stronger data augmentations:} Inspired by SimCLR~\cite{chen2020simple}, we select ColorJitter, GrayScale and GaussianBlur as extra data augmentations, and we apply them on Point DETR and baseline. The results are shown in Table~\ref{tab.ablations.augmentation}, demonstrate that with stronger augmentations to teacher model, the performance can be further improved.
% 18.2(-2.7)  23.7(-1.7)  29.3(-1.6)  32.9(-0.4)  34.8(-0.1)  36.3(0.9)
% \begin{figure}[t]
% \centering
% \includegraphics[width=.44\textwidth]{figures/mIoU_compare.pdf}
% \captionsetup{font=small}
% % \setcaptionwidth{.4\textwidth}
% \caption{Comparison in mIoUs between generated pseudo-boxes and ground truth boxes of FCOS$^\dagger$ and Point DETR}\label{fig.results.mIoU}\vspace{-0.3cm}
% \end{figure}
\paragraph{Absolute vs. Relative Regression:}\label{sec.ablations.regress} Our method use relative regression to predict object boxes. In Figure~\ref{fig.ltrb.vs.xywh}, we compare our relative regression with absolute regression used in DETR. Absolute regression incorrectly matches the point with the bounding box that does not correspond (\eg the green clock in Figure~\ref{fig.ltrb.vs.xywh}b) in some cases. Compared with absolute regression, relative regression has little mismatch problem between point and object box, we attribute it to its use of the prior knowledge: the point is \emph{in} the bounding box.

% In other words, box predictions is easily far away the corresponding point annotations using absolute regression. 

% Absolute regression tends to degenerate in some cases, \ie it regresses to a significantly simple bounding box, while ignoring its own bounding box which is hard to regress, \eg the green clock in Figure~\ref{fig.xywh}.

% Compared with absolute regression, relative regression has little mismatch problem between point and object box. In other words, box predictions is easily far away the corresponding point annotations using absolute regression. 

\paragraph{Point Annotations:} To evaluate the effectiveness of point annotations, we compare our approach with the method without point annotations. For a fair comparison, we use DETR as the method without point annotations. We apply a self-training framework (following ~\cite{sohn2020simple}) on DETR directly. We train DETR with only fully labeled images first, and then generate pseudo-boxes for weakly labeled images \emph{without} point annotations. To remove duplicate boxes, we use a threshold $\tau=0.7$ which results in the best box predictions on weakly labeled images. For the generated pseudo-object boxes, they do not have the one-to-one correspondence with point annotations. Thus, it is impractical to calculate the $mIoU$ between generated boxes and ground truth boxes. To make comparison available, we use standard COCO metrics instead of $mIoU$, as shown in Table~\ref{tab.ablations.pseudo.quality}. Point DETR performs on par with DETR on mAP and outperforms DETR by a large margin in the recall. Specifically, Point DETR achieves over 10 points of improvements in various AR metrics (\eg AR$_s$, AR$_m$, AR$_l$, AR$_{100}$) and its AP is comparable with DETR (26.8 vs. 26.9). Though Point DETR is 3.3 points $AP_{75}$ lower than DETR, which is possibly explained by that high $\tau$ screens out low-quality boxes and remains high-quality boxes, the higher recall of Point DETR can offset this bad influence. 

% We also notice a drop in AP$_{75}$, we conjecture it is because the opponent (\ie DETR) screens out the low-quality  boxes by $\tau$, which would result in higher quality of boxes with a large reduction in the number of boxes.

Additionally, we set the classification score of pseudo-boxes generated by DETR to a constant value like 0.5, which is consistent with our method. In this setting, the performance of DETR drops by a large margin and performs much worse than our method. This highlights that with point annotations, our method does not suffer from the quality of classification score.

\begin{figure}[t]
\centering
\includegraphics[width=.45\textwidth]{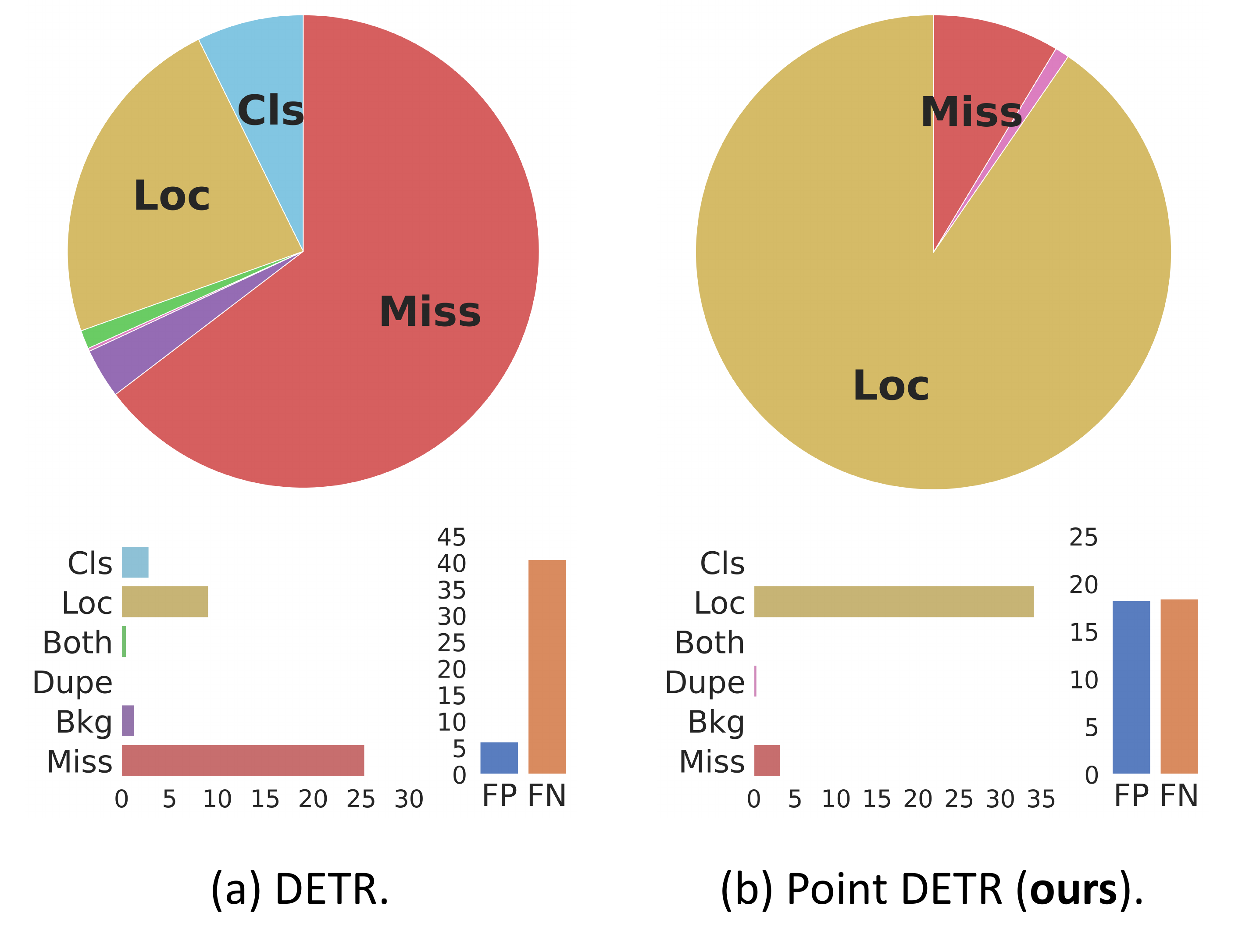}
% \begin{subfigure}[b]{.45\linewidth}
% \centering
% \includegraphics[width=1.\textwidth]{figures/error_analysis_detr.pdf}
% \captionsetup{font=small}
% \setcaptionwidth{.9\textwidth}
% \vspace{-0.3cm}
% \caption{DETR.}\label{fig.errors.detr0.7}
% \end{subfigure}
% \begin{subfigure}[b]{.45\linewidth}
% \centering
% \includegraphics[width=1.\textwidth]{figures/error_analysis_pointdetr.pdf}
% \captionsetup{font=small}
% \setcaptionwidth{.9\textwidth}
% \vspace{-0.4cm}
% \caption{Point DETR (\textbf{ours}).}\label{fig.errors.pointdetr}
% \end{subfigure}
\caption{Diagnosing the errors of generated pseudo-boxes by TIDE~\cite{bolya2020tide}. Different error types: \textbf{Cls}: localized correctly but classified incorrectly, \textbf{Loc}: classified correctly but localized incorrectly, \textbf{Both}: both cls and loc error, \textbf{Dupe}: duplicate detection error, \textbf{Bkg}: detected background as foreground, \textbf{Miss}: missed ground truth error.}
\label{fig.errors}
% \vspace{-0.2cm}
\end{figure}

We also analyze the errors of the generated boxes by TIDE~\cite{bolya2020tide} in Figure~\ref{fig.errors}. Missed ground truths is the largest issue for DETR, while it does not affect the performance of Point DETR greatly. This is explained by that with point annotations, Point DETR does not miss objects like DETR. In addition, unlike DETR, location error is the main challenge of Point DETR. Also, Point DETR also has duplicate detection errors. This is caused by those point annotations residing in multiple bounding boxes that would predict object boxes for wrong ground truths, which results in a ground truth that has multiple box predictions.

% This maybe result from that Point DETR (only regression) can not benefit from multi-task learning like DETR.
% \vspace{-0.cm}
\section{Conclusion}
In this work, we verify the effectiveness of point annotations in the weakly semi-supervised detection task. We also show that the power of point annotations is hindered by existing detectors. In order to solve this, we propose the Point DETR which applies a point encoder to the point annotations to establish the one-to-one correspondence between point annotations and objects.  Our approach is simple and implemented easily.  We demonstrate its efficacy by the extensive experimental analysis showing that it achieves state-of-the-art performance.

\paragraph{Acknowledgments}
This work is supported by National Key R\&D Program of China (2020AAA0105200).

\clearpage

{\small
\bibliographystyle{ieee_fullname}
\bibliography{egbib}
}

\end{document}